\documentclass[runningheads]{llncs}
\usepackage{graphicx}

\usepackage{float}
\usepackage{amsmath}
\usepackage{booktabs}
\usepackage{amssymb}

\usepackage{pifont}
\newcommand{\cmark}{\ding{51}}
\newcommand{\xmark}{\ding{55}}
\usepackage{wasysym}
\usepackage{tikz,pgfplots}

\usetikzlibrary{arrows}

\usetikzlibrary{decorations.pathmorphing}
\usetikzlibrary{positioning}
\usepackage{xcolor}
\usepackage{xspace}
\usepackage{cite}
\usepackage{caption}
\usepackage{subcaption}
\usepackage[hidelinks]{hyperref}

\newcommand{\langProp}[1]{\ensuremath{\mathcal{L}(#1)}\xspace}	
\newcommand{\true}{\ensuremath{1}}
\newcommand{\false}{\ensuremath{0}}					
\newcommand{\interpretationsProp}[1]{\ensuremath{\Omega(#1)}}
	
\newcommand{\modelSet}[1]{\ensuremath{\mathsf{Mod}(#1)}}

\newcommand{\LTL}{LTL\xspace}

\newcommand{\LTLf}{LTL$_\text{f}$\xspace}
\newcommand{\LTLff}{LTL$_{\text{ff}}$\xspace}

\newcommand{\kb}{\ensuremath{\mathcal{K}}}

\newcommand{\atoms}{\ensuremath{\mathsf{At}}}
\newcommand{\allkbs}{\ensuremath{\mathbb{K}}}
\newcommand{\allltl}{\ensuremath{\mathbb{L}}}

\newcommand{\MI}{\textsf{MI}}
\newcommand{\Free}{\textsf{Free}}

\newcommand{\inc}{\ensuremath{\mathcal{I}}}
\newcommand{\posRealInf}{\ensuremath{\mathbb{R}^{\infty}_{\geq 0}}}

\newcommand{\x}{\textbf{X}\xspace}

\newcommand{\un}{\textbf{U}\xspace}

\newcommand{\g}{\textbf{G}\xspace}
\newcommand{\f}{\textbf{F}\xspace}

\newcommand{\valueP}[1]{\textsc{Value}$_{#1}$}
\newcommand{\upperP}[1]{\textsc{Upper}$_{#1}$}
\newcommand{\lowerP}[1]{\textsc{Lower}$_{#1}$}
\newcommand{\exactP}[1]{\textsc{Exact}$_{#1}$}

\newcommand{\incdrastic}{\ensuremath{\mathcal{I}_{d}}}
\newcommand{\incmi}{\ensuremath{\mathcal{I}_{\mathsf{MI}}}}

\newcommand{\incc}{\ensuremath{\mathcal{I}_{c}}}
\newcommand{\incr}{\ensuremath{\mathcal{I}_{r}}}

\newcommand{\incp}{\ensuremath{\mathcal{I}_{p}}}

\newcommand{\incmv}{\ensuremath{\mathcal{I}_{mv}}}
\newcommand{\incat}{\ensuremath{\mathcal{I}_{at}}}

\newcommand{\incltld}{\ensuremath{\mathcal{I}^{LTL}_d}}
\newcommand{\incltl}{\ensuremath{\mathcal{I}^{LTL}_c}}

\begin{document}
\title{Measuring Inconsistency in\\Declarative Process Specifications\thanks{This work has been accepted for publication at the 20th International Conference on Business Process Management (BPM 2022). This work has been partially supported by the Deutsche Forschungsgemeinschaft (Grant DE 1983/9-1).}}
%
%
\author{Carl Corea\inst{1} \and
John Grant\inst{2} \and
Matthias Thimm\inst{3}}
\authorrunning{C. Corea et al.}
%
\institute{
Institute for IS Research, University of Koblenz-Landau, Koblenz, Germany\\
\email{ccorea@uni-koblenz.de}\\
\and
University of Maryland, College Park, USA\\
\email{grant@cs.umd.edu}\\
 \and
Artificial Intelligence Group, University of Hagen, Hagen,  Germany\\
\email{matthias.thimm@fernuni-hagen.de}
}
\maketitle              
\begin{abstract}
We address the problem of measuring inconsistency in 
declarative process specifications, with an emphasis on 
linear temporal logic on fixed traces (\LTLff). As we will show, existing inconsistency measures for classical logic cannot provide a meaningful assessment of inconsistency in LTL in general, as they cannot adequately handle the temporal operators. We therefore propose a novel paraconsistent semantics 
as a framework for inconsistency measurement. We then present two new inconsistency measures based on these semantics and show that they satisfy important desirable properties. We show how these measures can be applied to declarative process models and investigate the computational complexity of the introduced approach.

\keywords{Inconsistency Measurement 
\and LTL \and Declare.}
\end{abstract}
%
%
%

%
%
\section{Introduction}\label{sec:introduction}
Linear temporal logic (LTL) is an important logic for specifying the (temporal) behavior of business processes in the form of \emph{declarative process specifications} \cite{DBLP:conf/focs/Pnueli77,cecconi2020temporal}. The underlying idea is that time is represented as a linear sequence of states $T=(t_0,...,t_m)$, where $t_0$ is the designated starting point. At every state, some statements may be true. 
Temporal operators specify properties that must hold over the sequence of states.
For example, the 
operator \x (\emph{next}) means that a certain formula holds at the next state. Likewise, the 
operator \g (\emph{globally}) means that a certain formula will hold for all following states. 
Note that we interpret over finite traces, i.e., we consider a linear temporal logic over finite traces (\LTLf) \cite{DBLP:conf/focs/Pnueli77,DBLP:conf/ijcai/GiacomoV13}.

Traditionally, model checking has been used to verify that a particular model---that is, the assignment of truth values for statements over the time sequence---satisfies the requirements.
However, a problem in this use case arises if the set of 
formulas is \emph{inconsistent}, i.\,e., contains contradictory specifications. In such a case, the set of specifications cannot be applied for its intended purpose of process verification. For example, consider the two sets of 
formulas $\kb_1$ and $\kb_2$ (we will formalize syntax and semantics later):
\begin{align*}
    \kb_1 = \{\x a, \x\neg a\} && \kb_2 = \{\g a, \g \neg a\}
\end{align*}
Both $\kb_1$ and $\kb_2$ are inconsistent, as they demand that both $a$ and $\neg a$ hold in (some) following state, which is unsatisfiable. This calls for the analysis of such inconsistencies, to provide insights for inconsistency resolution. 

In classical logic, all inconsistent sets are equally bad \cite{knight2002measuring}. However, considering again the two sets,
intuitively, $\kb_2$ is ``more" inconsistent than $\kb_1$: The inconsistency in $\kb_1$ only affects the next state, while the inconsistency in $\kb_2$ affects all following states. This is an important insight that could prove useful for debugging or re-modelling \LTLf specifications or \LTLf-based constraint sets in general such as Declare. While there have been some recent works that can \emph{identify} inconsistent sets in declarative process specifications \cite{diciccio:2017resolving,corea2021interactive,roveri:2022}, those works cannot look ``into" those sets or compare them. In this work, we therefore show how to distinguish the \emph{severity} of inconsistencies in \LTLf, specifically, \LTLff.

A scientific field geared towards the quantitative assessment of inconsistency in knowledge representation formalisms is \emph{inconsistency measurement} \cite{grant2018measuring,Thimm:2019d}, and therefore represents a good candidate for this endeavour. 
Inconsistency measurement studies measures that aim to assess a \emph{degree} of inconsistency with a numerical value. The intuition here is that a higher value represents a higher degree of inconsistency. Such measures can provide valuable insights for debugging inconsistent 
specifications, e.\,g., to determine whether certain sets of formulas are more inconsistent than others. As we will show, existing measures are currently not geared towards \LTLf and temporal operators, and therefore cannot provide a meaningful analysis. Therefore, the main goal of this paper is to develop a new approach for measuring inconsistency in linear temporal logic. To frame this problem, we introduce a variant of \LTLf, which we coin linear temporal logic on \emph{fixed} traces \LTLff (cf. Section \ref{sec:ltlff}). 

Our contributions are as follows. We formalise the problem of measuring inconsistency in \LTLff and propose a rationality postulate that should be met by quantitative measures applied to this setting (Section \ref{sec:preliminaries}). We show that existing inconsistency measures do not satisfy this property, and propose an approach for measuring inconsistency 
based on a novel paraconsistent semantics for \LTLff (Section \ref{sec:mainSection}). 
We then show how our approach can be applied for measuring inconsistency in declarative process models (Section \ref{sec:declare}). For evaluation, we investigate the computational complexity of central aspects regarding inconsistency measurement in \LTLff (Section \ref{sec:complexity}). A conclusion is provided in Section \ref{sec:discussion}. Proofs for technical results can be found in the \hyperlink{target}{Appendix}.

\addtocounter{footnote}{-3}


%
%

\section{Preliminaries}\label{sec:preliminaries}
The traditional setting for inconsistency measurement is that of propositional logic.
For that, let $\atoms$ be some fixed propositional signature, i.\,e., a (possibly infinite) set of propositions, and let $\langProp{\atoms}$ be the corresponding propositional language constructed using the usual connectives $\wedge$ (\emph{conjunction}), $\vee$ (\emph{disjunction}), 
and $\neg$ (\emph{negation}). A literal is a proposition $p$ or negated proposition $\neg p$. 
\vspace{-.3cm}
\begin{definition}\label{def:kb}
	A knowledge base $\kb$ is a finite set of formulas $\kb\subseteq\langProp{\atoms}$. Let $\allkbs$ be the set of all knowledge bases.
\end{definition}
For a set of formulas $X$ we denote the set of propositions in $X$ by $\atoms(X)$.

Semantics for a propositional language is given by \emph{interpretations} where an interpretation $\omega$ on \atoms\ is a function $\omega:\atoms\rightarrow\{\false,\true\}$ (where $0$ stands for false and $1$ stands for true). Let $\interpretationsProp{\atoms}$ denote the set of all interpretations for \atoms. An interpretation $\omega$ \emph{satisfies} (or is a \emph{model} of) an atom $a\in\atoms$, denoted by $\omega\models a$, if and only if $\omega(a)=\true$. The satisfaction relation $\models$ is extended to formulas in the usual way. For $\Phi\subseteq\langProp{\atoms}$ we also define $\omega\models \Phi$ if and only if $\omega\models\phi$ for every $\phi\in\Phi$.
Furthermore, for every set of formulas $X$, the set of models is 
$\modelSet{X}=\{\omega\in \interpretationsProp{\atoms}\mid \omega\models X\}$. Define $X\models Y$ for (sets of) formulas $X$ and $Y$ if $\omega\models X$ implies $\omega\models Y$ for all $\omega$.

Let $\top$ denote any tautology and $\perp$ any contradiction. If $\modelSet{X}=\emptyset$ we write $X\models \perp$ and say that $X$ is \emph{inconsistent}. 

\subsection{Inconsistency Measurement}\label{sec:incMeasurement}
Inconsistency as defined above is a binary concept. To provide more fine-grained insights on inconsistency beyond such a binary classification, the field of inconsistency measurement \cite{Thimm:2019d} has evolved. 
The main objects of study in this field are \emph{inconsistency measures}, which are quantitative measures that assess the degree of inconsistency for a knowledge base $\kb$ with a non-negative numerical value. Intuitively, a higher value reflects a higher degree, or severity, of inconsistency. This can be useful for determining if one set of formulas is ``more" inconsistent than another. Let $\posRealInf$ be the set of non-negative real values including $\infty$. Then, an inconsistency measure is defined as follows.
\begin{definition}\label{def:inconsistenyMeasure}
	An inconsistency measure $\inc$ is any function $\inc:\allkbs\rightarrow \posRealInf$.
\end{definition}
To constrain the desired behavior of concrete inconsistency measures, several properties, called {\em rationality postulates}, have been proposed.
A well-agreed upon property is that of \emph{consistency}, which states that an inconsistency measure should return a value of 0 iff there is no inconsistency.
\begin{description}
	\item[\emph{Consistency} (\textsf{CO})] $\inc(\kb)=0$ if and only if $\kb$ is consistent.
\end{description}
Further important postulates introduced in \cite{hunter2006shapley} are \emph{monotony, dominance} and \emph{free-formula independence}, which we will define below. 
For that, we need some further notation.

First, a set $M\subseteq\kb$ is called a \emph{minimal inconsistent subset} (\textsf{MIS}) of $\kb$ if $M\models \perp$ and there is no $M'\subset M$ with $M'\models \perp$. Let $\MI(\kb)$ be the set of all \textsf{MIS}s of $\kb$. Second, a formula $\alpha\in\kb$ is called a \emph{free formula} if $\alpha\notin\bigcup\MI(\kb)$. Let $\Free(\kb)$ be the set of all free formulas of $\kb$. 

For the remainder of this section, let $\inc$ be an inconsistency measure, $\kb,\kb'\in\allkbs$, and $\alpha,\beta\in \langProp{\atoms}$. Then, the basic postulates from \cite{hunter2006shapley} are defined as follows.
\begin{description}
	\item[\emph{Monotony} (\textsf{MO})] If $\kb\subseteq \kb'$ then $\inc(\kb)\leq \inc(\kb')$.
	\item[\emph{Free-formula independence} (\textsf{IN})] If $\alpha\in\Free(\kb)$ then\\ $\inc(\kb)=\inc(\kb\setminus\{\alpha\})$.
	\item[\emph{Dominance} (\textsf{DO})] If $\alpha\not\models\perp$ and $\alpha\models\beta$ then $\inc(\kb\cup\{\alpha\})\geq\inc(\kb\cup\{\beta\})$.
\end{description}
\textsf{MO} states that adding formulas to the knowledge base cannot decrease the inconsistency value. \textsf{IN} means that removing free formulas from the knowledge base does not change the inconsistency value. \textsf{DO} consists of several cases, depending on the presence or absence of $\alpha$ or $\beta$ in $\kb$: the idea is that substituting a consistent formula $\alpha$ by a weaker formula $\beta$ cannot increase the inconsistency. 

Numerous inconsistency measures have been proposed (see \cite{Thimm:2019a} for a survey), many of which differ in regard to their compliance w.r.t. the introduced postulates. In this work, we will consider six measures as defined below. In order to define the \emph{contension measure} $\incc$ \cite{grant2011measuring} we need some additional background on Priest's three-valued semantics \cite{Priest:1979}. A three-valued interpretation is a function $\nu: \atoms \rightarrow \{0,1,\text{B}\}$, which assigns to every atom either $0$, $1$ or $\text{B}$, where $0$ and $1$ correspond to $false$ and $true$, respectively, and $\text{B}$ (standing for \emph{both}) denotes a conflict. 
Assuming the \emph{truth order} $\prec_T$ with $0\prec_T B \prec_T 1$, the function $\nu$ can be extended to arbitrary formulas as follows: $\nu(\alpha\wedge\beta) = \min_{\prec_T}(\nu(\alpha),\nu(\beta))$, $\nu(\alpha\vee\beta) = \max_{\prec_T}(\nu(\alpha),\nu(\beta))$, $\nu(\neg \alpha)=1$ if $\nu(\alpha)=0$, $\nu(\neg \alpha)=0$ if $\nu(\alpha)=1$, and $\nu(\neg \alpha)=B$ if $\nu(\alpha)=B$.
We say that an interpretation $\nu$ satisfies a formula $\alpha$, denoted by $\nu \models^3 \alpha$, iff $\nu(\alpha)=1$ or  $\nu(\alpha)= \text{B}$. 

We will now define the measures used in this work.

\begin{definition}
    Let the measures $\incdrastic$, $\incmi$, $\incp$, $\incr$, $\incc$, and $\incat$ be defined as follows:
    \begin{align*}
        \incdrastic(\kb) & = \left\{\begin{array}{cc}                                
				1                & \text{if~} \kb\models\perp                              \\
				0                & \text{otherwise}
				\end{array}\right.\\[1ex]
		\incmi(\kb)      & =|\MI(\kb)|  		\\[1ex]
		\incp(\kb)       & = |\bigcup_{M\in\MI(\kb)}M|                               \\[1ex]
		\incr(\kb)      & = \min\{|X| \mid X \subseteq \kb \text{~and } \kb \setminus X \not\models\perp\}\\[1ex]
		\incc(\kb)       & = \min\{|\nu^{-1}(B)\cap \atoms|\mid\nu\models^{3} \kb\}   \\[1ex]
		\incat(\kb)      & = |\bigcup_{M\in\MI(\kb)}\atoms(M)|
    \end{align*}
\end{definition}

A baseline approach is the drastic inconsistency measure $\incdrastic$ \cite{hunter2008measuring}, which only differentiates between inconsistent and consistent knowledge bases. The $\MI$-inconsistency measure $\incmi$ \cite{hunter2008measuring} counts the number of minimal inconsistent subsets. A similar version is the problematic inconsistency measure $\incp$ \cite{grant2011measuring}, which counts the number of distinct formulas appearing in any inconsistent subset. The repair measure $\incr$ counts the smallest number of formulas that must be removed in order to restore consistency. The contension measure $\incc$ \cite{grant2011measuring} quantifies inconsistency by seeking a three-valued interpretation that assigns $\text{B}$ to a minimal number of propositions. Finally, the $\incat$ measure counts the number of atoms in the non-free formulas.

We conclude this section with a small example illustrating the behavior of the considered inconsistency measures.
\begin{example}
Consider $\kb_3$, defined via
\begin{align*}
    \kb_3 &= \{a, \neg a, b, \neg b\wedge c \wedge d, \neg a \vee \neg b\}
\end{align*}
Then we have that
\begin{align*}
    \MI(\kb_3) &= \{\{a, \neg a\},\{b, \neg b \wedge c \wedge d\}, \{a, \neg a \vee \neg b, b\}\}
\end{align*}
Thus
\begin{align*}
    \incdrastic(\kb_{3})&=1 & \incmi(\kb_{3}) & = 3 & \incp(\kb_{3}) & = 5\\
    \incr(\kb_3)&=2     &\incc(\kb_{3})&=2   & \incat(\kb_3) & = 4
\end{align*}
\end{example}

The main focus of study in inconsistency measurement, and the introduced measures, has been on propositional logic. In this work, our aim is to apply inconsistency measures for linear time logic, which we introduce now.

\subsection{Linear Temporal Logic on Fixed Traces}\label{sec:ltlff}
In this work, we consider a specific variant of \LTLf that we coin linear temporal logic on \emph{fixed} traces (\LTLff). We consider a linear sequence  of states $t_0,\ldots,t_m$, where every $t_i$ is the state at instant $i$. We assume that $m>1$ to avoid the trivial case. Note that the difference with \LTLf---where interpretations can vary in their length as long as they are finite---is that we keep the length of this sequence finite and fixed across all interpretations. This variant of \LTLf is introduced mainly to discuss matters of inconsistency measurement, as here, the inconsistency value is computed in regard to a comparable length for all formulas. However, the ideas presented in the next sections can be extended to \LTLf \cite{DBLP:conf/aaai/FiondaG16} in a straightforward manner: In the unbounded case we can use a parameter N and then proceed as in the bounded case. This also means that $m$ must not necessarily be known or provided a priori, as a parameter N can be selected.

The syntax of \LTLff is the same as the syntax of \LTL and \LTLf \cite{DBLP:conf/ijcai/GiacomoV13}. Formulas are built from a set of propositional symbols $\atoms$ and are closed under the Boolean connectives, the unary operator \x (\emph{next}), and the binary operator \un (\emph{until}). Formally, any formula $\varphi$ of \LTLff is built using the grammar rule
\begin{align*}
    \varphi ::= a|(\neg\varphi)|(\varphi_1 \wedge \varphi_2)|(\varphi_1 \vee \varphi_2)|(\x\varphi)|(\varphi_1 \un \varphi_2).
\end{align*}
with $a\in\atoms$. Intuitively, $\x\varphi$ denotes that $\varphi$ will hold at the next state and $(\varphi_1 \un \varphi_2)$ denotes that $\varphi_1$ will hold until the state when $\varphi_2$ holds.
Let $d(\varphi)\in\mathbb{N}$ denote the maximal number of nested temporal operators in $\varphi$.\footnote{$d(\varphi)$ is inductively defined via $d(a)=0$ for $a\in \atoms$, $d(\neg \phi)=d(\phi)$, $d(\phi_1\wedge \phi_2)=d(\phi_1\vee \phi_2)=\max\{d(\phi_1),d(\phi_2)\}$, $d(\x\phi)=1+d(\phi)$, and $d(\phi_1 \un \phi_2)=1+\max\{d(\phi_1),d(\phi_2)\}$.}

From the basic operators, some useful abbreviations can be derived, including $\f\varphi$ (defined as $\top\un\varphi$), which denotes that $\varphi$ will hold (eventually) in the future and $\g\varphi$ (defined as $\neg\f\neg\varphi$), which denotes that $\varphi$ will hold for all following states. Again, let $\top$ be any tautology and $\bot$ any contradiction.

An \LTLff-interpretation $\hat{\omega}$ w.r.t.\ $\atoms$ is a function mapping each state and proposition to $0$ or $1$, meaning that $\hat{\omega}(t,a)=\true$ if proposition $a$ is assigned $\true$ (true) in state $t$.\footnote{Recall that we assume time of a fixed length $t_0,\ldots,t_m$ and interpretations only vary in what is true at each state.} Then the satisfaction of a formula $\phi$ by an interpretation $\hat{\omega}$, denoted by $\hat{\omega}\models \phi$, is defined via
\begin{align*}
    \hat{\omega} \models \phi \quad\Leftrightarrow\quad \hat{\omega}, t_0 \models \phi
\end{align*}
where $\hat{\omega}, t_i \models \phi$ for any interpretation $\hat{\omega}$ as above and for every $t_i\in\{t_0,…,t_m\}$ is inductively defined as follows:
\begin{align*}
    &\hat{\omega},t_i \models a \text{ iff } \hat{\omega}(t_i,a)=\true \text{~for~}a\in\atoms\\
    &\hat{\omega},t_i \models \neg\varphi \text{ iff } \hat{\omega},t_i \not\models \varphi\\
    &\hat{\omega},t_i \models \varphi_1 \wedge \varphi_2 \text{ iff } \hat{\omega},t_i\models \varphi_1 \text{ and } \hat{\omega},t_i\models \varphi_2\\
    &\hat{\omega},t_i \models \varphi_1 \vee \varphi_2 \text{ iff } \hat{\omega},t_i\models \varphi_1 \text{ or } \hat{\omega},t_i\models \varphi_2\\
    &\hat{\omega},t_i \models \x\varphi \text{ iff } i<m \text{~and~} \hat{\omega},t_{i+1} \models \varphi \\ 
    &\hat{\omega},t_i \models \varphi_1 \un \varphi_2 \text{ iff } \hat{\omega},t_j \models \varphi_2 \text{ for some } j\in\{i+1,\ldots,m\}\\ 
    &\qquad\quad\quad\quad\quad\text{ and } \hat{\omega},t_k \models \varphi_1 \text{ for all } k\in\{i,\ldots,j-1\}
\end{align*}
An interpretation $\hat{\omega}$ satisfies a set of formulas $K$ iff $\hat{\omega} \models \phi$ for all $\phi\in K$. A set $K$ is consistent iff there exists $\hat{\omega}$ such that $\hat{\omega}\models K$. Define $X\models Y$ for (sets of) formulas $X$ and $Y$ if $\hat{\omega}\models X$ implies $\hat{\omega}\models Y$ for all $\hat{\omega}$.

\subsection{Related Work and Contributions}
This work is related to consistency- and model checking in declarative process specifications, see e.g. \cite{new2,new3,new5}. In particular, our approach extends recent works \cite{diciccio:2017resolving,corea2021interactive,roveri:2022,new5} on the identification of inconsistent sets in declarative process specifications by allowing to look ``into" those sets and leverage inconsistency resolution with quantitative insights. For example, existing resolution approaches mainly try to minimize the \emph{number} of deleted formulas \cite{diciccio:2017resolving,corea2021interactive,maggi2011runtime}. This however completely leaves aside the semantics of those formulas or their impact on any corresponding process. Given this motivation, it is useful to consider also the degree to which certain formulas affect the following behavior, which is why we propose time sensitive inconsistency measures.

This paper is related to \cite{grant2021measuring} which presents several, what we call time sensitive, inconsistency measures for branching time logics (BTL). 
However, in this work we are able to avoid the complicated overload of branching time as the process specifications are provided in linear time logic. Using branching time logic adds a layer of complexity that is unnecessary when dealing with a linear time situation. Just to take one example, consider the set $\{\x a, \x\neg a\}$. In linear time logic this gives one inconsistency at the next state. But in the case of branching time logic what does $\x$ mean? There may be many ``next" states. If $\x$ means ``some next state" then the set is consistent because $a$ and $\neg a$ may hold in different next states. If $\x$ means ``all next states" then it is inconsistent but how inconsistent depends on the number of next states. We avoid such issues by dealing only with linear temporal logic. Note also that BTL takes a different view on time than \LTLf as studied in this paper and is therefore expressively incomparable (cf. \cite{vardi2001branching}). 

\section{Inconsistency Measurement in \LTLff}\label{sec:mainSection}
In this section, we address the issue of measuring inconsistency in \LTLff. As we will show, existing inconsistency measures cannot provide meaningful insights when dealing with temporal logic. Therefore, we develop a novel paraconsistent semantics as a framework for handling inconsistency 
and propose two concrete inconsistency measures for \LTLff.

\subsection{Motivation for Inconsistency Measures for \LTLff}\label{sec:motivation}

We recall the sets of \LTLff formulas $\kb_1$ and $\kb_2$:
\begin{align*}
    \kb_1 = \{\x a, \x\neg a\} && \kb_2 = \{\g a, \g \neg a\}
\end{align*}
The knowledge base $\kb_1$ states that $a$ is both true and false in the next state while $\kb_2$ states that $a$ is both true and false in all future states. Obviously, both knowledge bases are inconsistent. Yet, the inconsistencies are different in regard to the number of states they affect. 
For $\kb_1$ the number is $1$ and for $\kb_2$ the number is $m>1$. 
It would therefore be desirable for an inconsistency measure to take this information into account and assign $\kb_2$ a larger inconsistency value.

%

In order to capture \LTLff\ by the inconsistency measurement framework of Section~\ref{sec:incMeasurement}, from now on a knowledge base $\kb$ (Definition~\ref{def:kb}) will be a finite set of \LTLff formulas and $\allkbs$ is the set of all \LTLff knowledge bases. 
So we can apply the inconsistency measures for $\kb_1$ and $\kb_2$ in a straightforward manner.

\begin{example}\label{ex:currentMeasuresNotCorrect}
Consider $\kb_1$ and $\kb_2$. Then we have that
\begin{align*}
    \incdrastic(\kb_{1})&=1 & \incdrastic(\kb_{2})&=1\\
    \incmi(\kb_{1})&=1 & \incmi(\kb_{2})&=1\\
    \incp(\kb_{1})&=2 & \incp(\kb_{2})&=2\\
    \incr(\kb_{1})&=1 & \incr(\kb_{2})&=1\\
    \incc(\kb_{1})&=1 & \incc(\kb_{2})&=1\\
    \incat(\kb_{1})&=1 & \incat(\kb_{2})&=1
\end{align*}
\end{example}
Note that all six inconsistency measures give identical values for $\kb_1$ and $\kb_2$, because they, or for that matter, any other propositional logic inconsistency measure, cannot distinguish between $\x$ and $\g$. But intuitively $\kb_2$ is more inconsistent than $\kb_1$ because the inconsistency persists through all future states in $\kb_2$ as opposed to the single state in $\kb_1$. Thus, we believe that a proper inconsistency measure for \LTLff should distinguish between these operators. Therefore, we propose a new rationality postulate.

\begin{description}
	\item[\emph{Time Sensitivity} (\textsf{TS})] 
	For all formulas $\varphi$ of propositional logic, $\inc(\{\g\varphi,\g\neg\varphi\}) >  
	\inc(\{\x\varphi,\x\neg\varphi\})$. 
\end{description}
In other words, the number of affected states should be reflected in the inconsistency value, i.e., inconsistency measures for \LTLff should be time sensitive. 

\begin{proposition}\label{prop:tsViolation}
$\incdrastic, \incmi, \incp, \incr, \incc, \incat$ violate \textsf{TS}.
\end{proposition}
Following Proposition \ref{prop:tsViolation}, the existing measures that we have from propositional logic cannot capture the desired behavior. Therefore, we introduce a novel approach to measure inconsistency in \LTLff.

%
%
\subsection{A Paraconsistent Semantics for \LTLff}\label{sec:para}

Our first contribution towards measuring inconsistency in \LTLff is to define an \LTLff-variant of the three-valued semantics of \cite{Priest:1979}. By doing so, we not only develop a means to neatly express inconsistency measures for \LTLff, but also define a general applicable paraconsistent semantics for \LTLff.

A three-valued interpretation $\hat{\nu}$ for \LTLff is a function mapping each state and proposition to $0$, $1$ or $\text{B}$, that is, 
$\hat{\nu}: \{t_0, t_1, \ldots t_m\} \times \textsf{At} \rightarrow \{0, 1, \text{B}\}$ where as before $0$ and $1$ correspond to the classic logical false and true, respectively, and $\text{B}$ (standing for \emph{both}) denotes a conflict. We then assign
\begin{align*}
    \hat{\nu}(\phi) &= \hat{\nu}(t_0,\phi)
\end{align*}
where $\hat{\nu}(t_i,\phi)$, for any interpretation $\hat{\nu}$ as above and state $t_i\in\{t_0,…,t_m\}$, is inductively defined as follows: 
\begin{align*}
    \hat{\nu}(t_i,a) &= \hat{\nu}(t_i,a) \text{~for~}a\in\atoms\\
    \hat{\nu}(t_i,\neg \phi) & = \left\{\begin{array}{ll}
                                    1 & \text{if~} \hat{\nu}(t_i, \phi)=0\\
                                    0 & \text{if~} \hat{\nu}(t_i, \phi)=1\\
                                    \text{B} & \text{if~} \hat{\nu}(t_i, \phi)=\text{B}
                                \end{array}\right.\\
    \hat{\nu}(t_i,\varphi_1 \wedge \varphi_2) &= \left\{\begin{array}{ll}
                                    1 & \text{if~} \hat{\nu}(t_i, \varphi_1)=\hat{\nu}(t_i, \varphi_2)=1\\
                                    0 & \text{if~} \hat{\nu}(t_i, \varphi_1)=0 \text{~or~} \hat{\nu}(t_i, \varphi_2)=0\\
                                    \text{B}& \text{otherwise}
                                \end{array}\right.\\
    \hat{\nu}(t_i,\varphi_1 \vee \varphi_2) &= \left\{\begin{array}{ll}
                                    1 & \text{if~} \hat{\nu}(t_i, \varphi_1)=1 \text{~or~}\hat{\nu}(t_i, \varphi_2)=1\\
                                    0 & \text{if~} \hat{\nu}(t_i, \varphi_1)=\hat{\nu}(t_i, \varphi_2)=0\\
                                    \text{B} & \text{otherwise}
                                \end{array}\right.\\
\end{align*}
\begin{align*}
    \hat{\nu}(t_i,\x\varphi) &= \left\{\begin{array}{ll}
                                    \hat{\nu}(t_{i+1},\varphi) & \text{if~} i<m\\
                                    0 & \text{otherwise}
                                \end{array}\right.\\
    \hat{\nu}(t_i,\varphi_1 \un \varphi_2) &= \left\{\begin{array}{ll}
                                    1 & \text{if there is~} j\in\{i+1,\ldots,m\} \text{ with}\\
                                                & \hat{\nu}(t_j,\varphi_2)=\hat{\nu}(t_i,\varphi_1)=\ldots\\
                                                &=\hat{\nu}(t_{j-1},\varphi_1)=1\\
                                    \text{B} & \text{if there is~} j\in\{i+1,\ldots,m\} \text{ with}\\
                                                & \{\hat{\nu}(t_j,\varphi_2),\hat{\nu}(t_i,\varphi_1),\ldots,\\
                                                &\hat{\nu}(t_{j-1},\varphi_1)\}=\{1,\text{B}\}\\
                                    0 & \text{otherwise}
                                \end{array}\right.
\end{align*}
Some comments on the above definition are in order. First, note that the evaluation of the classical Boolean connectives is the same as for propositional three-valued semantics (see Section \ref{sec:incMeasurement}). Furthermore, the evaluation of $\x\phi$ is simply the truth value of $\phi$ at the next state, or, if there is no next state, $0$ (as for the classical semantics of \LTLff). The main new feature, however, is the three-valued evaluation of a formula 
of the form
$\varphi_1 \un \varphi_2$. This formula evaluates to $1$ as in the classical case, i.\,e., if $\phi_2$ evaluates to $1$ in some future state and $\phi_1$ evaluates to $1$ in between. We evaluate $\varphi_1 \un \varphi_2$ to $\text{B}$ if $\phi_2$ evaluates to $1$ or $\text{B}$ in some future state and $\phi_1$ evaluates to $1$ or $\text{B}$ in between (and at least one of these evaluations must be to $\text{B}$). Finally, $\varphi_1 \un \varphi_2$ evaluates to $0$ otherwise, i.e., if either $\phi_2$ always evaluates to $0$ in the future or in-between $\varphi_1$ evaluates at least once to $0$. 

A three-valued \LTLff interpretation $\hat{\nu}$ satisfies a formula $\phi$, denoted by $\hat{\nu}\models^3\phi$, iff $\hat{\nu}(\phi, t_0)\in\{1, \text{B}\}$. A three-valued interpretation $\hat{\nu}$ satisfies a set of formulas $\kb$ iff $\hat{\nu} \models^3 \phi$ for all $\phi\in \kb$.
\begin{example}
    Let $\atoms=\{a,b\}$ and assume $m=2$. Consider the knowledge base $\kb$ defined via
    \begin{align*}
        \kb & = \{\x \neg a, a\un b  \}
    \end{align*}
    and the three-valued interpretation $\hat{\nu}$ defined via
    \begin{align*}
        \hat{\nu}(t_0,a)&=1 & \hat{\nu}(t_0,b) &= 0\\
        \hat{\nu}(t_1,a)&=\text{B} & \hat{\nu}(t_1,b) &= 0\\
        \hat{\nu}(t_2,a)&=0 & \hat{\nu}(t_2,b) &= 1
    \end{align*}
    Then we have $\hat{\nu}(t_0,a\un b)=\text{B}$ as $b$ evaluates to $1$ in $t_2$ and $a$ evaluates to $\text{B}$ in $t_1$. Moreover, we have $\hat{\nu}(t_0,\x \neg a)=\text{B}$ and therefore $\hat{\nu}\models^3\kb$.
\end{example}
Define $X\models^3 Y$ for formulas $X$ and $Y$ if $\hat{\nu}\models X$ implies $\hat{\nu}\models Y$ for all $\hat{\nu}$.

In the propositional logic case, $\models^3$ is a faithful extension of $\models$, meaning that $\omega\models \phi$ if and only if $\omega\models^3 \phi$ for every two-valued interpretation $\omega$ and every $\phi$. Our \LTLff extension of the three-valued semantics enjoys the same property (note that every two-valued interpretation is a also a three-valued interpretation that does not use the value $\text{B}$).

\begin{proposition}\label{th:faithful}
    For every (two-valued) \LTLff interpretation $\hat{\omega}$ and \LTLff formula $\phi$, $\hat{\omega}\models \phi$ if and only if $\hat{\omega}\models^3 \phi$.
\end{proposition}
The three-valued semantics of \cite{Priest:1979} has another nice property in propositional logic, namely the non-existence of inconsistency: every propositional formula is trivially satisfiable by the interpretation that assigns \text{B} to all propositions. 
In general, an \LTLff formula may become unsatisfiable w.r.t.\ to the three-valued semantics if it affects a state ''beyond`` $t_m$. However, for other formulas we obtain the following result regarding universal satisfiability.
\begin{proposition}
    For any \LTLff formula $\phi$ with $d(\phi)\leq m$ there is $\hat{\nu}$ with $\hat{\nu}\models^3 \phi$.
\end{proposition}
The semantics presented in this section allows for inconsistency-tolerant reasoning in \LTLff\ (and it can straightforwardly be adapted for \LTLf\ and \LTL). This provides a useful tool for the usual application scenarios of temporal logics, such as model checking and verification. While it may be worthwhile to investigate this aspect in more depth, in the remainder of this work we will focus on the application of this semantics for inconsistency measurement and postpone that endeavour to future work.

\subsection{Time Sensitive Inconsistency Measures for \LTLff}\label{sec:measurement}
We will now exploit our three-valued semantics for \LTLff to define some new inconsistency measures. We do this similarly as for propositional logic by assessing the amount of usage of the paraconsistent truth value $\text{B}$ in models of an \LTLff knowledge base $\kb$ but refine it by two different levels of granularity. This yields two new inconsistency measures.

Our first approach measures the number of states affected by inconsistency. For any three-valued interpretation $\hat{\nu}$, define
\begin{align*}
    \textsf{AffectedStates}(\hat{\nu}) & = \{t\mid \exists a:\hat{\nu}(t,a)=\text{B}\}
\end{align*}
In other words, $\textsf{AffectedStates}(\hat{\nu})$ is the set of states where $\hat{\nu}$ assigns $\text{B}$ to at least one proposition. We can define an inconsistency measure by considering those 3-valued models of the knowledge base that affect the minimal number of states.
\begin{definition}[LTL time measure]
Let $\kb$ be a set of formulas. Then, the LTL time measure is defined via
\begin{align*}
    \incltld(\kb) &= \min_{\hat{\nu}\models^3 \kb} |\textsf{AffectedStates}(\hat{\nu})|
\end{align*}
if there is $\hat{\nu}$ with $\hat{\nu}\models^3 \kb$ and $\incltld(\kb)=\infty$ otherwise.
\end{definition}
This measure counts the number of states for which the knowledge base is inconsistent. It is, in fact, the extension of the drastic measure, $\incdrastic$, in that for each state it adds $1$ if there is an inconsistency and $0$ otherwise. This measure can be used to distinguish the knowledge bases $\kb_1$ and $\kb_2$, i.e., it is time sensitive.
\begin{example}\label{ex:incltldFirstExample}
We recall the knowledge bases $\kb_1 = \{\x a, \x\neg a\}$ and $\kb_2 = \{\g a, \g\neg a\}$. Then we have
\begin{align*}
    \incltld(\kb_1)=1 && \incltld(\kb_2)=m
\end{align*}
\end{example}
As an example where there is no $\hat{\nu}$ s.t. $\hat{\nu}\models^3 \kb$, consider the formula $\x\x\x a$. This formula cannot be satisfied for $m=2$, so $\incltld$ would return $\infty$ here.

Example \ref{ex:incltldFirstExample} shows that the proposed measure $\incltld$ can already provide meaningful insights for measuring inconsistency in LTL. But a potential limitation is that it can only distinguish inconsistency in individual states in a binary manner. For example, $\incltld$ cannot distinguish the knowledge base $\kb_4=\{\x a, \x\neg a, \x b, \x \neg b\}$ from $\kb_1$ because all inconsistencies occur at one state, namely $t_1$. For this reason we believe it is useful to be able to look inside states for inconsistency. In order to do so,
given a three-valued interpretation $\hat{\nu}$, define
\begin{align*}
    \textsf{Conflictbase}(\hat{\nu}) & = \{(t,a)\mid \hat{\nu}(t,a)=\text{B}\}
\end{align*}
Then, define the LTL contension measure as follows.
\begin{definition}[LTL contension measure]
Let $\kb$ be a set of formulas and 
\begin{align*}
    \incltl(\kb) &= \min_{\hat{\nu}\models^3 \kb} |\textsf{Conflictbase}(\hat{\nu})|
\end{align*}
if there is $\hat{\nu}$ with $\hat{\nu}\models^3 \kb$ and $\incltl(\kb)=\infty$ otherwise.
\end{definition}

$\incltl$ seeks an interpretation that assigns $\text{B}$ to a minimal number of propositions individually over all the states and uses this number for the inconsistency measure. This is an extension of $\incltld$, and for that matter, of $\incc$ as it calculates $\incltl$ for each state and sums the numbers obtained this way. 
\begin{example}\label{ex:myExample2}
We recall the knowledge bases $\kb_1 = \{\x a, \x\neg a\}$, $\kb_4 = \{\x a, \x\neg a,$ $\x b, \x \neg b\}$, and consider $\kb_5 = \{\g a, \g\neg a, \g b, \g \neg b\}$. If $m=3$, then we have
\begin{align*}
    \incltld(\kb_1) = 1 && \incltld(\kb_4) = 1 && \incltld(\kb_5) = 3\\
    \incltl(\kb_1) = 1 && \incltl(\kb_4) = 2 && \incltl(\kb_5) = 6
\end{align*}
\end{example}


As can be seen in Example \ref{ex:myExample2}, the two inconsistency measures proposed in this work can, contrary to previously existing measures, be used to provide meaningful insights into inconsistency in linear temporal logic, i.e., they are in fact time sensitive. As the two measures have a different granularity in regard to time, selecting which of the two to use depends on the intended use case. 



Intuitively, it would be possible to devise further time-sensitive inconsistency measures for \LTLff. We will however leave this discussion for future work.
Importantly, the aim of this paper is to show that traditional inconsistency measures cannot be plausibly applied to temporal logics, and to present means for time sensitive inconsistency measurement. In this regard, the measures proposed in this work can be used as a baseline for measuring inconsistency in LTL. Also, they (broadly) satisfy other desirable properties and can therefore be seen as strictly better (w.r.t. the considered postulates) than their propositional logic ``counterpart", i.e., $\incdrastic$ for $\incltld$, respectively $\incc$ for $\incltl$ . The results of this section are summarized in Table \ref{tab:complianceWithRationalityPostulates}. Proofs can be found in the \hyperlink{target}{Appendix}. 

\begin{table}
    \centering
    \begin{tabular}{|c|c|c|c|c|c|}\hline
    $\inc$	        &\textsf{CO}  &\textsf{MO} &\textsf{IN} &\textsf{DO}    & \textsf{TS}	\\\hline
    $\incdrastic$   &\cmark&\cmark&\cmark&\cmark&\xmark\\
    $\incmi$        &\cmark&\cmark&\cmark&\xmark&\xmark\\
    $\incp$         &\cmark&\cmark&\cmark&\xmark&\xmark\\
    $\incr$         &\cmark&\cmark&\cmark&\xmark&\xmark\\
    $\incc$         &\cmark&\cmark&\xmark&\cmark&\xmark\\
    $\incat$        &\cmark&\xmark&\xmark&\xmark&\xmark\\
    $\incltld$      &\cmark&\cmark&\cmark&\cmark&\cmark\\
    $\incltl$       &\cmark&\cmark&\xmark&\cmark&\cmark\\
    \hline
    \end{tabular}
    \caption{Compliance of inconsistency measures with rationality postulates.}
    \label{tab:complianceWithRationalityPostulates}
\end{table}

Note that only the measures we introduced satisfy \textsf{TS}. Note also that $\incltl$ does not satisfy \textsf{IN} due to the problem of iceberg inconsistencies, cf. the provided \hyperlink{target}{proofs}.

\section{Application to Declarative Process Models}\label{sec:declare}
A common application scenario for \LTLf is that of declarative process models \cite{pesic2007declare}, which are sets of (LTL-based) constraints. For such declarative process models, the issue of inconsistency is equally as problematic, as any inconsistencies between the constraints make the declarative process model unsatisfiable.

There have been a number of works addressing the issue of inconsistency in declarative process models \cite{diciccio:2017resolving,maggi2011runtime,corea2021interactive}. However, those works mainly look at whether a process model is inconsistent at all (in a binary manner), or try to identify sets of inconsistent constraints. Those works can however not look ``into" those sets or assess their severity. For this use case, our proposed approach can be extended to declarative process models as follows.


\subsection{Inconsistency Measurement in Declarative Process Models}

A declarative process model consists of a set of constraints. Typically, these constraints are constructed using predefined templates, i.\,e., predicates, that are specified relative to a set of propositions (e.\,g., company activities).

\begin{definition}[Declarative Process Model]
A declarative process model is a tuple $\textbf{M} = (\textbf{A}, \textbf{T}, \textbf{C})$, where $\textbf{A}$ is a set of propositions, $\textbf{T}$ is a set of constraint types, and $\textbf{C}$ is the set of constraints, which instantiate the template elements in $\textbf{T}$ with activities in $\textbf{A}$.\footnote{For readability, we will denote declarative process models as a set of constraints ($\textbf{C}$)}
\end{definition}

In this work, we consider the declarative modelling language Declare \cite{pesic2007declare}, which offers a set of ``standard" templates. 
We will use a selection of templates shown in Table \ref{tab:my_declare_templates}. We refer the reader to \cite{diciccio:2017resolving} for an overview of other Declare template types and corresponding semantics.
\begin{table}[H]
    \centering
    \begin{tabular}{l|l}
    \textbf{Template} & \textbf{\LTLff Semantics}\\
    \hline
        \text{Init}(a)              & a\\
        \text{End}(a)               & $\g(a\vee\f a)$\\
        \text{Response}(a,b)        & $\g(a\rightarrow \f b)$ \\
        \text{NotResponse}(a,b)     & $\g(a\rightarrow \neg\f b)$ \\
        \text{ChainResponse}(a,b)   & $\g(a\rightarrow \x b)$ \\
        \text{NotChainResponse}(a,b)\text{ } & $\g(a\rightarrow \neg\x b)$ \\
        \text{AtLeast}(a,\emph{n})  & $\f(a \wedge \x(\text{atLeast}(a, \text{\emph{n}-1}))), \text{atLeast}(a, 1) = a \vee \f(a)$ \\
        \text{AtMost}(a,\emph{n})   & $\g(\neg a \vee \x(\text{atMost}(a, \text{\emph{n}}-1))), \text{atMost}(a, 0) = \g(\neg a)$ \\
    \end{tabular}
    \caption{\LTLff Semantics for a selection of Declare templates.}
    \label{tab:my_declare_templates}
\end{table}

By rewriting the constraints of a declarative process model into \LTLff formulas, our approach for measuring inconsistency in \LTLff can be applied to Declare in a straightforward manner.

\begin{example}\label{ex:declareEx1}
Consider the sets of constraints $C_a$ and $C_b$, defined via
\begin{align*}
    C_a &= \{\textsc{Init}(a), \textsc{Response}(a,b),\textsc{NotResponse}(a,b)\}\\
    &(\Leftrightarrow \{a, \g(a\rightarrow\f b), \g(a\rightarrow \neg\f b)\})\\ 
    C_b &= \{\textsc{Init}(a), \textsc{Response}(a,b), \textsc{NotResponse}(a,b),\\
    &\qquad\textsc{Response}(a,c), \textsc{NotResponse}(a,c)\}\\ 
    &(\Leftrightarrow \{a, \g(a\rightarrow\f b), \g(a\rightarrow \neg\f b), \g(a\rightarrow\f c), \g(a\rightarrow \neg\f c)\})
\end{align*}
then we have that 
$\incltl(C_a)=1$ and $\incltl(C_b)=2$.
\end{example}
Due to the recursive definition of some ``existence" constraints (cf. Table \ref{tab:my_declare_templates}), note that also inconsistencies concerned with cardinalities can be assessed correctly.
\begin{example}
Consider $C_c = \{\textsc{AtMost}(a,1)$,
  $\textsc{AtLeast}(a,2)\}$ and \\$C_d = \{\textsc{AtMost}(a,1)$,
  $\textsc{AtLeast}(a,100)\}$, then $\incltld(C_c)<\incltld(C_d)$.
\end{example}
As a border case, note that any inconsistency referring to a point in time beyond the assumed sequence of states will return a value of $\infty$ per definition, as we cannot assess any error that leaves the boundaries of our logical framework.
\begin{example}
Let $C_e = \{\textsc{End}(a), \textsc{ChainResponse}(a,b)\}$, then $\incltld(C_e)=\infty$.
\end{example}

These examples show that
our approach can provide detailed insights on the severity of inconsistency in declarative process models. Such insights can prove useful for prioritizing or re-modelling different issues of the process specification. 
In this context, it seems intuitive that conflicts affecting only the next state ($\x$) 
should be considered as less severe than conflicts affecting multiple following states ($\g$), i.\,e., for any \LTLff formula $\varphi$, $\inc(\{\g\varphi,\g\neg\varphi\}) >  
	\inc(\{\x\varphi,\x\neg\varphi\})$. In this regard, there are 
	still open questions on how to distinguish the operators $\x$ and $\f$, in particular: for an \LTLff formula $\varphi$, what is the relation between $\inc(\{\x\varphi,\neg\x\varphi\})$ and $\inc(\{\f\varphi,\neg\f\varphi\})$? We address this question in the following.

\subsection{On Potentially Inconsistent States}

Consider the following sets of constraints $C_m$ and $C_n$, defined via
\begin{footnotesize}
\begin{align*}
    &C_m = & &C_n =  \\
    &\{\textsc{Init}(a)\Leftrightarrow a, &  &\{\textsc{Init}(a),\\
    &\textsc{Response}(a,b)\Leftrightarrow \g(a\rightarrow\f b), & &\textsc{ChainResponse}(a,b)\Leftrightarrow \g(a\rightarrow\x b),\\
    &\textsc{NotResponse}(a,b)\Leftrightarrow \g(a\rightarrow\neg\f b)\} & &\textsc{NotChainResponse}(a,b)\Leftrightarrow \g(a\rightarrow\neg\x b)\}
\end{align*}
\end{footnotesize}
Both sets are inconsistent, as they demand that $b$ should and should not follow. 
However, the point in time at which the actual inconsistency can occur is different. Naturally, one question arises: which inconsistency is more severe? Or are they equally severe? We encourage the reader to come up with an own answer to this question at this point before we continue with our view on this matter.

\begin{figure}[H]
    \tikzset{vertex/.style = {shape=circle,draw,minimum size=1.8em}}
    \tikzset{vertexWhite/.style = {shape=circle,draw=white,minimum size=1.8em}}
    \tikzset{vertexDashed/.style = {shape=circle,draw,dashed,minimum size=1.8em,line width=0.1mm}}
    \tikzset{vertexThick/.style = {shape=circle,draw,minimum size=1.8em,line width=0.4mm}}
    \tikzset{vertexGray/.style = {shape=circle,draw=gray,minimum size=1.8em}}
    \tikzset{edge/.style = {->}}  
        
    \centering
    \begin{subfigure}{.49\textwidth}
        \centering
        \begin{tikzpicture}
        \node[vertex] (a) {\emph{a}};
        \node[vertexDashed,right of = a, node distance = 0.45in] (b) {?};
        \node[vertexWhite,right of = b, node distance = 0.45in] (c) {...};
        \node[vertexDashed,right of = c, node distance = 0.45in] (d) {?};

        \draw[edge] (a) to (b);
        \draw[edge] (b) to (c);
        \draw[edge] (c) to (d);
        \end{tikzpicture}
        \caption{Potentially inconsistent states for $C_m$}
    \end{subfigure}
    \begin{subfigure}{.49\textwidth}
        \centering
        \begin{tikzpicture}
        
        \node[vertex] (a) {\emph{a}};
        \node[vertexThick,right of = a, node distance = 0.45in] (b) {\lightning};
        \node[vertexGray,right of = b, node distance = 0.45in] (c) {\hphantom{A}};
        \node[vertexWhite,right of = c, node distance = 0.45in] (d) {...};

        \draw[edge] (a) to (b);
        \draw[edge] (b) to (c);
        \draw[edge] (c) to (d);
        \end{tikzpicture}
        \caption{Certainty of inconsistency for $C_n$}
    \end{subfigure}
    
    \caption{Visualization of the (un)certainty of where the inconsistency may occur for $C_m$ and $C_n$.}
    \label{fig:certainty_of_inconsistency_example}
\end{figure}
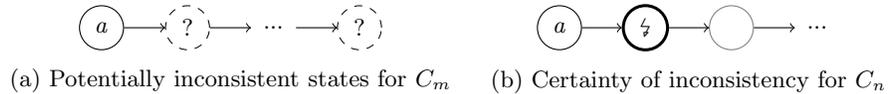

Using the measures introduced in this work, the absolute number of affected states is 1 in both cases. So regarding the minimal number of affected states, the inconsistencies are equally severe. However, the certainty of where the inconsistency can occur at is clearly different, as visualized in Figure \ref{fig:certainty_of_inconsistency_example}. 

\vspace{-.4cm}
In $C_m$, there are $m$ different possible states to which a minimal interpretation could assign the truth value $\text{B}$ to the proposition $b$, whereas the inconsistency can only occur in exactly 1 state for $C_n$. This could entail different severities for the inconsistencies, depending on the viewpoint:


Consider a running process which is in state $t_0$. For $C_m$, it is unclear when the inconsistency will occur. For $C_n$, it is directly known that the next state is inconsistent. Recovery mechanisms for such cases are well known \cite{ly2013framework}, e.\,g., it would be possible to just skip the next state and continue with a consistent process. This is not possible for $C_m$ without skipping all following states until the end of the process. So one might argue that the inconsistency in $C_m$ is more severe. However, for $C_n$, this also means there is in fact no possible continuation as the process is in a dead-end state, thus, $C_n$ needs to be attended to more urgently (So one might as well argue that the inconsistency in $C_n$ is more severe).

In the field of inconsistency measurement, the dominance property states that substituting a consistent formula by a weaker formula cannot increase the inconsistency value \cite{hunter2006shapley}. However, when moving from $C_m$ to $C_n$ or vice-versa, we both replace one constraint with a stronger one and the other with a weaker one (every \textsc{ChainResponse} is also a \textsc{Response} but every \textsc{NotResponse} is a \textsc{NotChainResponse}). So the dominance property is not applicable here and the question remains which inconsistency is more severe. In this work, we will not give a definitive answer to this question and leave this discussion for future work. However, based on the two possible views given above, we will argue that they are, in fact, different. It would therefore be desirable to be able to distinguish the inconsistency in $C_m$ and $C_n$. Here, the introduced contension concept can be adapted to quantify the certainty of \emph{when} the inconsistency will occur. 

The introduced measures quantify inconsistency by seeking an interpretation that assigns $\text{B}$ to a minimal number of states. We denote the set of all such interpretations that assign $\text{B}$ to a minimal number of states (at least to one) as
\begin{align*}
    \hat{V}_{\textsf{min}}^{B>0}(\kb) &=
    \{\hat{\nu}\models^3\kb : |\textsf{AffectedStates}(\hat{\nu})|>0 \wedge |\textsf{AffectedStates}(\hat{\nu})| = \incltld(\kb) \}
\end{align*}
Every such (minimal) interpretation also encodes which exact states are affected by the inconsistency. For $C_n$, only one state is necessarily affected (cf. Fig \ref{fig:certainty_of_inconsistency_example} (b)), thus, there exists only one minimal interpretation. For $C_m$, there are $m$ different interpretations that are all equally minimal in terms of how many states are affected. So the number of minimal interpretations relates to the number of distinct (sets of) states that can potentially be affected. 

\begin{definition}[Number of Minimal Interpretations]
Let $\kb$ be a set of formulas. Then, define the number of minimal interpretations via
\begin{align*}
    \textsf{\#minInterpretations}(\kb) = |\hat{V}_{\textsf{min}}^{B>0}(\kb)|
\end{align*}
\end{definition}


\begin{example}
We recall $C_m$ and $C_n$. Then we have that $\textsf{\#minInterpretations}(C_m)=m$ and $\textsf{\#minInterpretations}(C_n)=1$ as expected (cf. the above discussion)
\end{example}

Importantly, the function $\textsf{\#minInterpretations}$ is \underline{not} an inconsistency measure, i.\,e., a higher value does not indicate a higher degree of inconsistency. It therefore also does not matter where the inconsistency in $C_m$ eventually triggers. The value merely expresses the ``certainty" of knowing where the conflict can occur at. The semantics of which is worse depends on the use case.

%
%
\section{Computational Complexity}\label{sec:complexity}
We conclude with an investigation of computational complexity in measuring inconsistency in LTL. We assume familiarity with computational complexity, see \cite{Papadimitriou:1994a} for an introduction. Proofs can be found in the \hyperlink{target}{Appendix}.

Note that deciding satisfiability is \textsf{PSPACE}-complete for \LTLf \cite{DBLP:conf/ijcai/GiacomoV13} and also intractable for many variants of \LTLf \cite{DBLP:conf/aaai/FiondaG16}. For our variant \LTLff, as $m$ is fixed, we get \textsf{NP}-completeness (think for example of a non-deterministic algorithm that guesses $\hat{\omega}$ and verifies (in polynomial time) that $\hat{\omega}\models \phi$).
\begin{theorem}\label{th:satisfy}
    Deciding whether a formula $\phi$ is satisfiable in \LTLff is \textsf{NP}-complete.
\end{theorem}

If the parameter $m$ is given in unary, the complexity result holds as it is. However, if $m$ is given in binary then the complexity will likely increase (in the membership proof, we need to guess an interpretation and if m is given in binary, that interpretation may be exponential in the size of the input). 

We continue with an investigation of the computational complexity of measuring inconsistency in LTL. For this, let $\allltl$ denote the set of all \LTLff knowledge bases.
Following \cite{Thimm:2019a}, we consider the following computational problems:
\begin{tabbing}
	\exactP{\inc}  \qquad\= \textbf{Input}: \quad \=$\kb\in\allltl$, $x\in\posRealInf$ \\
		\> \textbf{Output}: \> \textsc{true} iff $\inc(\kb)=x$	\\[1ex]
	\upperP{\inc} \> \textbf{Input}: \quad \>$\kb\in\allltl$, $x\in\posRealInf$ \\
		\> \textbf{Output}: \> \textsc{true} iff $\inc(\kb)\leq x$	\\[1ex]
	\lowerP{\inc} \> \textbf{Input}: \quad \>$\kb\in\allltl$, $x\in\posRealInf\setminus\{0\}$ \\
		\> \textbf{Output}: \> \textsc{true} iff $\inc(\kb)\geq x$  \\[1ex]
	\valueP{\inc} \> \textbf{Input}: \>$\kb\in\allltl$ \\
		\> \textbf{Output}: \> The value of $\inc(\kb)$	
\end{tabbing}

For \upperP{\inc}, the same general non-deterministic algorithm 
can be applied.
\begin{theorem}\label{th:complexity}
    \upperP{\incltld} and \upperP{\incltl} are \textsf{NP}-complete.
\end{theorem}
Using the results in \cite{Thimm:2019a} we also get the following results for the other problems.
\vspace{-.2cm}
\begin{corollary}
\lowerP{\incltld} and \lowerP{\incltl} are \textsf{coNP}-complete. \exactP{\incltld} and \exactP{\incltl} are in \textsf{DP}. \valueP{\incltld} and \valueP{\incltl} are in $\textsf{FP}^{\textsf{NP}[\log n]}$.
\end{corollary}

In regard to the algorithmic implementation of our approach, a general approach of SAT encodings can be used. Corollary 1 gives a straightforward implementation for an algorithm to compute the measures by combining binary search with iterative calls to a SAT solver using an encoding of the problem Upper (see proof of Corollary 1). This encoding would be based on a SAT encoding for \LTLff satisfiability, which is straightforward.

%
%
\section{Conclusion}\label{sec:discussion}
In this work, we have presented an approach for measuring the severity of inconsistencies in declarative process specifications, in particular those based on linear temporal logic. In this regard, we introduced a paraconsistent semantics for \LTLff and developed two inconsistency measures. This provides useful insights for debugging or re-modelling \LTLf specifications, e.\,g., by allowing to compare or prioritize different inconsistencies. 
Here, our approach extends recent works \cite{diciccio:2017resolving,corea2021interactive,roveri:2022} on the identification of inconsistent sets in declarative process specifications by allowing a look ``into" those sets.

In future work, we aim to investigate the application of our approach to other languages such as GSM or DCR. Note that this is however not trivial, as the process models there might not be represented as orthogonal formulas. As a further limitation of our work, the current approach treats time as discrete time steps where any number of activities (within the bounds of the constraints) are allowed to occur at the same time. Real processes may however contain activities that take real time and may not be parallelizable because of resource constraints. As a result, a logically equivalent inconsistency may weigh more than another. In future work, we aim to address this issue with data-aware versions of \LTLff.

\vspace{-.25cm}
\bibliographystyle{splncs04}
\bibliography{main}


\section*{\hypertarget{target}{Appendix}: Proofs for Technical Results}\label{sec:appendix}
\setcounter{proposition}{1}
\begin{proposition}\label{th:faithful}
    For every (two-valued) \LTLff interpretation $\hat{\omega}$ and \LTLff formula $\phi$, $\hat{\omega}\models \phi$ if and only if $\hat{\omega}\models^3 \phi$.
    \begin{proof}
        Let $\hat{\omega}$ be any two-valued \LTLff interpretation.
        First observe that using three-valued semantics, $\hat{\omega}(t_i,\phi)\neq \text{B}$ for every $\phi$ (this can be easily verified by structural induction since no atom has the value $\text{B}$).
        We prove now the more general statement that $\hat{\omega},t_i\models \phi$ if and only if $\hat{\omega}(t_i, \phi)=1$ for any $t_i$ by induction on the structure of $\phi$:
                \begin{itemize}
                    \item $\phi=a$ for $a\in \atoms$: From $\hat{\omega},t_i\models \phi$ it follows $\hat{\omega}(t_i,a)=\true$ via classical semantics, which is equivalent to $\hat{\omega}(t_i,a)=1$ via 3-valued semantics.
                    \item $\phi=\neg \psi$: We have $\hat{\omega},t_i\models \neg\psi$ iff $\hat{\omega},t_i\not\models \psi$. By the induction hypothesis, this is equivalent to $\hat{\omega}(t_i,\psi)=0$. This is equivalent to $\hat{\omega}(t_1,\neg \psi)=1$.
                    \item $\phi=\phi_1\wedge \phi_2$: We have that $\hat{\omega},t_i\models \phi_1\wedge \phi_2$ is equivalent to $\hat{\omega},t_i\models \phi_1$ and $\hat{\omega},t_i\models \phi_2$. By the induction hypothesis, this is equivalent to $\hat{\omega}(t_i,\phi_1)=1$ and $\hat{\omega}(t_i\phi_2)=1$. This is equivalent to $\hat{\omega}(t_i,\phi)=1$.
                    \item $\phi=\phi_1\vee \phi_2$: We have that $\hat{\omega},t_i\models \phi_1\vee \phi_2$ is equivalent to $\hat{\omega},t_i\models \phi_1$ or $\hat{\omega},t_i\models \phi_2$. Without loss of generality assume $\hat{\omega},t_i\models \phi_1$. By the induction hypothesis, this is equivalent to $\hat{\omega}(t_i,\phi_1)=1$ which implies $\hat{\omega}(t_i,\phi)=1$. This other direction is analogous.
                    \item $\phi=\x \psi$: Observe first that for $i\geq m$ both $\hat{\omega},t_i\not\models \phi$ and $\hat{\omega}(t_i, \phi)=0$ hold. So assume that $i< m$. Then we have $\hat{\omega},t_i\models \x\psi$ equivalent to $\hat{\omega},t_{i+1}\models \psi$. By the induction hypothesis, this is equivalent to $\hat{\omega}(t_{i+1}, \psi)=1$ and $\hat{\omega}(t_{i}, \x\psi)=1$.
                     \item $\phi=\varphi_1 \un \varphi_2$: Assume that $\hat{\omega},t_i\models \varphi_1 \un \varphi_2$. Then there is some $j\in\{i+1,\ldots,m\}$ with $\hat{\omega},t_j\models \phi_2$ and $\hat{\omega},t_k\models \phi_1$ for all $k\in \{i,\ldots,j-1\}$. By the induction hypothesis, this amounts to $\hat{\omega}(t_j,\phi_2)=1$ and $\hat{\omega}(t_k,\phi_1)=1$ for all $k\in \{i,\ldots,j-1\}$. From this follows $\hat{\omega}(t_i,\phi_1\un\phi_2)=1$. The other direction is analogous.
                \end{itemize}
    \end{proof}
\end{proposition}

\setcounter{proposition}{2}
\begin{proposition}
    For every \LTLff formula $\phi$ with $d(\phi)\leq m$ there is a $\hat{\nu}$ with $\hat{\nu}\models^3 \phi$.
    \begin{proof}
        Let $\hat{\nu}_b$ be the three-valued interpretation defined via $\hat{\nu}_b(t_i,a)=\text{B}$ for all $t_i$ and $a$. We prove the more general statement that $\hat{\nu}_b(t_i,\phi)=\text{B}$ for any $t_i$ and $d(\phi)\leq m-i$ by induction on the structure of $\phi$:
        \begin{itemize}
            \item $\phi=a$ for $a\in \atoms$: $\hat{\nu}_b(t_i,a)=\text{B}$ holds by definition.
            \item $\phi=\neg \psi$: $d(\phi)\leq m-i$ implies that $d(\psi)\leq m-i$ as well. By the induction hypothesis, $\hat{\nu}_b(t_i,\psi)=\text{B}$ and therefore $\hat{\nu}_b(t_i,\neg\psi)=\text{B}$ as well.
            \item $\phi=\phi_1\wedge \phi_2$: $d(\phi)\leq m-i$ implies that $d(\phi_1)\leq m-i$ and $d(\phi_2)\leq m-i$ as well. By the induction hypothesis, $\hat{\nu}_b(t_i,\phi_1)=\hat{\nu}_b(t_i,\phi_2)=\text{B}$ and therefore $\hat{\nu}_b(t_i,\phi_1\wedge\phi_2)=\text{B}$.
            \item $\phi=\phi_1\vee \phi_2$: $d(\phi)\leq m-i$ implies that $d(\phi_1)\leq m-i$ and $d(\phi_2)\leq m-i$ as well. By the induction hypothesis, $\hat{\nu}_b(t_i,\phi_1)=\hat{\nu}_b(t_i,\phi_2)=\text{B}$ and therefore $\hat{\nu}_b(t_i,\phi_1\vee\phi_2)=\text{B}$.
            \item $\phi=\x \psi$: $d(\phi)\leq m-i$ implies $d(\psi)\leq m-(i+1)$. By the induction hypothesis $\hat{\nu}_b(t_{i+1},\psi)=\text{B}$ which directly gives $\hat{\nu}_b(t_i,\x\psi)=\text{B}$.
            \item $\phi=\varphi_1 \un \varphi_2$: $d(\phi)\leq m-i$ implies that $d(\phi_1)\leq m-(i+1)$ and $d(\phi_2)\leq m-(i+1)$ as well. So, by the induction hypothesis $\hat{\nu}_b(t_{i+1},\phi_2)=\text{B}$ and therefore $\hat{\nu}_b(t_i,\varphi_1 \un \varphi_2)=\text{B}$. 
        \end{itemize}
    \end{proof}
\end{proposition}



\begin{table}
    \centering
    \begin{tabular}{|c|c|c|c|c|c|}\hline
    $\inc$	        &\textsf{CO}  &\textsf{MO} &\textsf{IN} &\textsf{DO}    & \textsf{TS}	\\\hline
    $\incdrastic$   &\cmark&\cmark&\cmark&\cmark&\xmark\\
    $\incmi$        &\cmark&\cmark&\cmark&\xmark&\xmark\\
    $\incp$         &\cmark&\cmark&\cmark&\xmark&\xmark\\
    $\incr$         &\cmark&\cmark&\cmark&\xmark&\xmark\\
    $\incc$         &\cmark&\cmark&\xmark&\cmark&\xmark\\
    $\incat$        &\cmark&\xmark&\xmark&\xmark&\xmark\\
    $\incltld$      &\cmark&\cmark&\cmark&\cmark&\cmark\\
    $\incltl$       &\cmark&\cmark&\xmark&\cmark&\cmark\\
    \hline
    \end{tabular}
    \caption{Compliance of the inconsistency measures with the considered rationality postulates.}
    \label{tab:complianceWithRationalityPostulates}
\end{table}

\begin{proposition}\label{prop:compliance}
The compliance of the inconsistency measures $\incdrastic, \incmi, \incp, \incr, \incc, \incat, \incltld$ and $\incltl$ with the postulates \textsf{CO}, \textsf{MO}, \textsf{IN}, \textsf{DO} and \textsf{TS} is as shown in Table \ref{tab:complianceWithRationalityPostulates}.

\begin{proof}
    The proofs for $\incdrastic,\incmi,\incp,\incr, \incc,\incat$ can be found in \cite{Thimm:2018}.\footnote{$\incr$ is equivalent to $\inc_{dalal}^{hit}$ from that work. Also, the proofs for $\incat$ are analogous to $\incmv$.} In the process we corrected one result in \cite{Thimm:2018}, based on an earlier error. Actually, $\incc$ does not satisfy \textsf{IN} because of iceberg inconsistencies (see \cite{debona2017iceberg}). For example, consider a knowledge base $\kb_c = \{a \wedge \neg a \wedge b, \neg b\}$. Then $\neg b$ is free, but $\incc(\kb_c) = 2$ and $\incc(\kb_c\setminus \{\neg b\}) = 1$. Also, for those measures \textsf{TS} follows from Example 2.

    We now consider the remaining measures $\incltld$ and $\incltl$ in turn. Recall that the definition of consistency for \LTLf requires the existence of a 2-valued interpretation $\hat{\nu}$ such that $\hat{\nu} \models \kb$. Using such an interpretation we obtain 
    $|\textsf{AffectedStates}(\hat\nu)|=|\textsf{ConflictBase}(\hat\nu)|=0$. If $\kb$ is inconsistent, \text{B} is assigned to at least one proposition, so these values are positive.  

    We now start with $\incltld$. For this, let $\kb, \kb'$ be knowledge bases and $\alpha, \beta$ be two formulas of \LTLff. \textsf{CO} follows directly from the definition of consistency. For \textsf{MO}, if $\kb$ is consistent, then $\kb\cup\kb'$ is either consistent or inconsistent. In both cases $\textsf{AffectedStates}(\kb\cup\kb') \geq \textsf{AffectedStates}(\kb)$. If $\kb$ is inconsistent, so is $\kb\cup\kb'$ and again $\incltld(\kb)\leq\incltld(\kb\cup\kb')$. For \textsf{IN}, observe that for any free formula $\alpha$, there exists an interpretation that maps $\alpha$ to $1$; thus, free formulas cannot affect the size of \textsf{AffectedStates}. In turn, if $\alpha$ is a free formula, for any interpretation $\hat\nu$ that assigns $\text{B}$ to a minimal number of propositions we have that $|\textsf{AffectedStates}(\hat\nu)|$ is the same for $\kb$ and ($\kb\setminus\alpha$). For \textsf{DO}, observe from \cite{hunter2010measure} that if $\alpha\models\beta$, then
    $\{\hat{\nu}\mid \hat{\nu}\models^3 (\kb\cup\{\alpha\})\}\subseteq \{\hat{\nu}\mid \hat{\nu}\models^3 (\kb\cup\{\beta)\})\}$. Therefore $\min_{\hat{\nu}\models^3 (\kb\cup\{\alpha\})} |\textsf{AffectedStates}(\hat{\nu})|\geq \min_{\hat{\nu}\models^3 (\kb\cup\{\beta\})} |\textsf{AffectedStates}(\hat{\nu})|$. 
    For \textsf{TS} let $\kb = \{\x \varphi, \x\neg \varphi\}$ and $\kb' = \{\g \varphi, \g \neg \varphi\}$. Then 
    there is an interpretation satisfying $\kb$ that only assigns \text{B} to state $t_1$. But for $\kb'$ every interpretation must assign \text{B} to $m$ states. Therefore $\min_{\hat{\nu}\models^3 \kb} |\textsf{AffectedStates}(\hat{\nu})|<\min_{\hat{\nu}\models^3 \kb'} |\textsf{AffectedStates}(\hat{\nu})|$.
        
    The proofs for $\incltl$ are analogous to $\incltld$ except that $\incltl$ counts the number of inconsistencies for each state also. $\textsf{VO}$ and $\textsf{MO}$ are immediate as before. For $\textsf{IN}$ the counterexample for propositional logic can be used. $\textsf{DO}$ follows as for $\incltld$. The only difference in the proof for $\textsf{TS}$ is that now it is possible to give a formula $\varphi$ for which $\incltl(\kb) >1$ but then $\incltl(\kb') = m \times \incltl(\kb) > \incltl(\kb)$.
\end{proof}
\end{proposition}

\setcounter{theorem}{0}
\begin{theorem}
    Deciding whether a formula $\phi$ is satisfiable in \LTLff is \textsf{NP}-complete.
    \begin{proof}
        For \textsf{NP}-membership consider the following non-deterministic algorithm. Since $m$ is fixed, we can guess an \LTLff interpretation $\hat{\omega}$ and verify (in polynomial time) that $\hat{\omega}\models \phi$. 
        
        \textsf{NP}-hardness follows directly from the fact that satisfiability in \LTLff subsumes propositional satisfiability (as $\phi$ can be a purely propositional formula).
    \end{proof}
\end{theorem}

\begin{theorem}\label{th:complexity}
    \upperP{\incltld} and \upperP{\incltl} are \textsf{NP}-complete.
    \begin{proof}
        First, observe that given a 3-valued interpretation $\hat{\nu}$, evaluating $\hat{\nu}(t_i,\phi)$ for any $t_i$ and $\phi$, as well as determining $\textsf{AffectedStates}(\hat{\nu})$ and $\textsf{Conflictbase}(\hat{\nu})$, can be done in polynomial time.
        
        For \textsf{NP}-membership consider the following non-deterministic algorithm. Given $\kb$ and $x\in\posRealInf$, we non-deterministically guess an interpretation $\hat{\nu}$ and verify $\hat{\nu}\models^3 \kb$ and $|\textsf{AffectedStates}(\hat{\nu})|\leq x$ (or $|\textsf{Conflictbase}(\hat{\nu})|\leq x$).
        
        For \textsf{NP}-hardness, observe that for given $\kb$ and $x=0$, both problems \upperP{\incltld} and \upperP{\incltl} are equivalent to the problem of deciding whether there is a two-valued interpretation that satisfies $\kb$. Due to Proposition~\ref{th:faithful} this problem is equivalent to the classical satisfiability problem in \LTLff. Therefore, we can reduce the classical satisfiability problem of propositional logic to the problem \upperP{\incltld} (or \upperP{\incltl}) with $x=0$, which shows the \textsf{NP}-hardness of the latter.
        \end{proof}
\end{theorem}

\setcounter{corollary}{0}
\begin{corollary}
\lowerP{\incltld} and \lowerP{\incltl} are \textsf{coNP}-complete. \exactP{\incltld} and \exactP{\incltl} are in \textsf{DP}. \valueP{\incltld} and \valueP{\incltl} are in $\textsf{FP}^{\textsf{NP}[\log n]}$.
\begin{proof}
    The \textsf{coNP}-completeness of \lowerP{\incltld} and \lowerP{\incltl} and the membership of \exactP{\incltld} and \exactP{\incltl} in \textsf{DP} follow from Lemma~6 in \cite{Thimm:2019a}, Proposition~\ref{th:complexity}, and the fact that the measures $\incltl$ and $\incltld$ are well-serializable, cf. Definition~22 in \cite{Thimm:2019a}, due to their range being equal to $\{0,\ldots,m,\infty\}$ (for $\incltld$) and $\{0,\ldots,m|\atoms|,\infty\}$ (for $\incltl$). The membership of \valueP{\incltld} and \valueP{\incltl} in $\textsf{FP}^{\textsf{NP}[\log n]}$ follow likewise from Lemma~4 in \cite{Thimm:2019a} and Proposition~\ref{th:complexity}.
\end{proof}
\end{corollary}
\end{document}